\pdfoutput=1
\documentclass{article}
\usepackage[main, final]{neurips_2026}
\makeatletter
\renewcommand{\@notice}{}
\makeatother
\usepackage{amsmath,amsfonts,bm}

\def\eqref#1{equation~\ref{#1}}
\def\1{\bm{1}}

\DeclareMathAlphabet{\mathsfit}{\encodingdefault}{\sfdefault}{m}{sl}
\SetMathAlphabet{\mathsfit}{bold}{\encodingdefault}{\sfdefault}{bx}{n}

\usepackage[utf8]{inputenc}
\usepackage[T1]{fontenc}
\usepackage{hyperref}
\usepackage{url}
\usepackage{booktabs}
\usepackage{amsfonts}
\usepackage{nicefrac}
\usepackage{microtype}
\usepackage{xcolor}
\usepackage{graphicx}
\usepackage{amsmath}
\usepackage{amssymb}
\usepackage{adjustbox}
\usepackage{algorithm}
\usepackage{algpseudocode}
\usepackage{float}
\usepackage{longtable}
\usepackage{makecell}
\title{Contamination, Prior, or Evidence? Decomposing and Training Evidence Use in Whole-Slide Vision–Language Models}

\author{%
  \textbf{Wenhao Zhang}\textsuperscript{1} \quad
  \textbf{Zhongliang Zhou}\textsuperscript{2} \quad
  \textbf{Shiyuan Zhang}\textsuperscript{1} \quad
  \textbf{Yiqing Yang}\textsuperscript{1} \\
  \textbf{Pinqiao Wang}\textsuperscript{1} \quad
  \textbf{Lehan Yang}\textsuperscript{1} \quad
  \textbf{Hanyin Wang}\textsuperscript{2} \quad
  \textbf{John Kang}\textsuperscript{2} \quad
  \textbf{Sheng Li}\textsuperscript{1}
  \\
  \normalfont\textsuperscript{1}University of Virginia, Charlottesville \\
  \textsuperscript{2}Merck \& Co., Inc.
}

\hypersetup{
  pdftitle={Contamination, Prior, or Evidence? Decomposing and Training Evidence Use in Whole-Slide Vision–Language Models},
  pdfauthor={Wenhao Zhang, Zhongliang Zhou, Shiyuan Zhang, Yiqing Yang, Pinqiao Wang, Lehan Yang, Hanyin Wang, John Kang, Sheng Li}
}

\begin{document}

\maketitle

\begin{abstract}

Pathology vision-language models (VLMs) are conventionally evaluated by accuracy, but accuracy alone does not measure evidence use: it may conflate dataset contamination, prior knowledge, and image evidence. In a motivating study of lymph-node metastasis prediction, we found that most public pathology VLMs showed minimal differences when changing from feeding the models with whole-slide images, an annotated lesion, or no image at all. To better understand the specific features leveraged by these models, this paper presents two contributions aimed at disentangling these factors. First, we present CleanSlide, a TCGA-based VQA benchmark designed to eliminate image- and question-side contamination. It contains 149K audited multiple-choice questions over 9,985 slides, with patient- and tissue-source-disjoint splits. Every question is audited for option shortcuts, stem leakage, cross-split duplication, and blind solvability. Second, we propose Pair-DPO, a preference loss over counterfactual slide pairs from the same question and source. By controlling for shared confounding factors, Pair-DPO cancels out the question-attributable signal and leaves image evidence as the source of preference. Specifically, each pair consists of two real slides with opposite, verified findings, introducing neither editing artifacts nor unverified labels for diffuse or graded features such as invasion, necrosis, and tumor grade. Experiments show that our method gains 15.29\% from image evidence on the CleanSlide, compared with 2.81\% for the best published model. On the external CPTAC and BCNB cohorts, our method achieves accuracies of 57.6\% and 59.0\%, outperforming all other evaluated models by 9.7\% and 3.4\%, respectively. We will release the benchmark and code.
\end{abstract}

\section{Introduction}

\begin{figure}
    \centering
    \includegraphics[width=0.9\linewidth]{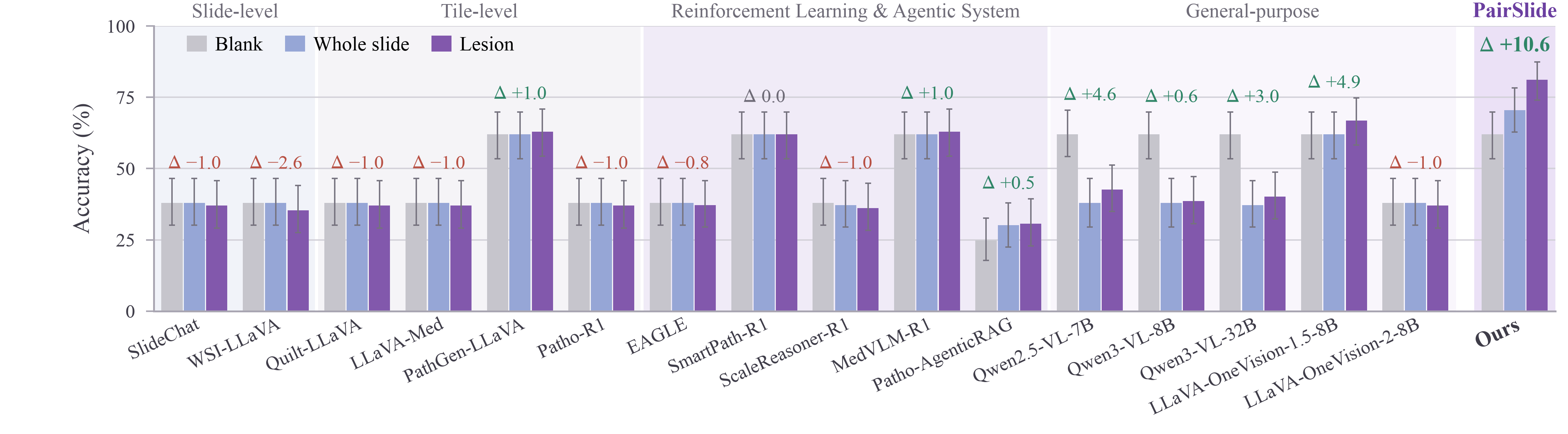}
    \caption{CAMELYON16 motivating study (n=129 slides; accuracy, \%, with 95\% bootstrap CIs). We evaluate 17 models across three evidence conditions—blank, whole slide, and lesion and test whether performance improves as increasingly relevant evidence is provided. $\Delta$ denotes the accuracy difference between whole-slide and lesion conditions.}
    \label{fig:whole-vs-lesion}
\end{figure}

Whole-slide images are the primary diagnostic record in histopathology, and
a single slide may hold billions of pixels of tissue information. Pathology vision--language models (VLMs)
that read such slides and answer clinical questions in free text are now
appearing at a steady pace \citep{slidechat,wsillava,cpathomni,pathchat}. Fundamentally, a pathologist’s diagnosis of a digital slide relies on visual examination. Therefore, we test whether VLMs actually use image evidence with a controlled intervention on the held-out CAMELYON16 cohort \citep{camelyon} (Figure~\ref{fig:whole-vs-lesion}). For the same metastasis-related question, each model is evaluated with a blank image, the full WSI, and the human-annotated lesion region. If a model truly leverages visual context, its performance should improve when transitioning from the blank baseline to the WSI, and increase further when focused on the precise lesion. Contrary to this expectation, most models output identical predictions across all three conditions, often repeating their diagnosis even when the clinical image is replaced with a blank canvas. Importantly, this failure is not limited to weaker architectures. It persists across a diverse spectrum of state-of-the-art systems: instruction-tuned tile-level models, slide-level architectures, reinforcement-learning-tuned pathology reasoners, and advanced agentic pipelines.

If diagnostic accuracy does not stem from visual evidence within the slide, what drives it? We decompose performance into three sources. The first is \textit{visual evidence}, where the model directly analyzes the slide and formulates its diagnostic answer based on the visual features present. The second is \textit{dataset contamination}, which encompasses exposure to patient- or site-level slides during training~\citep{zhang2026auditingdataleakagewholeslide}, as well as exposure to evaluation questions derived from the same medical report corpus used to mine the training data---a critical confounder that current slide-level benchmarks fail to control. The third is \textit{prior knowledge}, where the model derives its prediction solely from the text prompt; this linguistic prior is exceptionally strong in clinical contexts, where predicting the most statistically common diagnosis yields a high baseline rate of correct guesses. Consequently, a benchmark that reports aggregate accuracy alone conflates these three distinct sources, masking a model's true reliance on visual data.

Separating these three sources requires two critical components currently absent from standard practice: metrics that control for contamination while exposing the text-only prior, and training signal that incentivizes the model to let visual evidence dictate its decision. In natural image domains, visual evidence utilization is typically evaluated using blind baselines, contrast sets, and vision-indispensable benchmarks, and is subsequently improved through techniques like contrastive decoding and preference optimization~\citep{contrastsets,vqacp,mmstar,vcd,mdpo,povid,rlhfv}. However, these paradigms rest on three assumptions that do not hold in pathology, namely that the image can be edited to create a counterfactual sample, that the linguistic prior is a nuisance to be eliminated, and that the question primarily concerns simple object presence. Pathology violates all three conditions. Gigapixel whole-slide images are processed as thousands of distinct tile features, rendering faithful counterfactual image editing computationally and biologically impractical. Furthermore, clinical priors reflect real-world disease epidemiology and should be systematically measured rather than discarded. Finally, clinical questions target complex, graded histopathological findings such as microvascular invasion, tumor necrosis, and histopathological grade, rather than mere object detection. Consequently, pathology demands a unique alternative. Because diagnostic slides are natively paired with verified pathology reports alongside patient and tissue-source identifiers, we can construct natural counterfactual samples without image editing by leveraging a slide from a different patient with the same cancer type but an opposing, verified clinical finding.

Leveraging these resources, we first introduce the CleanSlide benchmark, comprising 148,654 multiple-choice questions across 9,985 whole-slide images spanning 32 cancer types curated exclusively from The Cancer Genome Atlas (TCGA). Unlike benchmarks scraped from web repositories, textbooks, or educational videos, TCGA provides the detailed provenance necessary to identify and eliminate overlap with pretraining corpora, thereby enabling a rigorous evaluation of genuine visual evidence utilization. Second, we present Pair-DPO, a preference optimization framework that exploits a fundamental property of pathology reports: standardized diagnostic findings such as lymphovascular invasion, necrosis, and margin status recur across patients with verified positive and negative ground truth. For instance, consider the query, \textit{``Is necrosis present in the invasive ductal carcinoma?''}, evaluated on two breast cancer slides whose reports record necrosis as \textit{Present} and \textit{Absent}, respectively. Pair-DPO optimizes the model to prefer \textit{Present} for the former and \textit{Absent} for the latter. Because the paired slides share the identical query, cancer type, and originating site, non-visual confounders are naturally controlled within each pair, allowing the objective function to directly reward predictions driven by histopathological tissue features. Our primary contributions are threefold:

\noindent \textbf{CleanSlide} comprises 9,985 TCGA slides and 149K multiple-choice questions reviewed by pathologists. It uses patient- and site-disjoint splits and audits every question for option shortcuts, stem leakage, cross-split duplication, and blind solvability. Our evaluation shows that existing pathology VLMs rely predominantly on prior knowledge rather than slide-based visual evidence.

\noindent \textbf{Pair-DPO} turns the standardized findings of pathology reports into real counterfactual slide pairs. The two slides of a pair share the question and the cancer type, so the prior cancels and the loss can only be lowered by reading the slide,
without editing any image or subtracting any prior. We also demonstrate that the loss can transfer beyond pathology to the natural image domain.

\noindent \textbf{PairSlide} is a new slide-level VLM model trained with Pair-DPO. On CleanSlide, it reaches the best accuracy in our comparison and raises counterfactual sensitivity without losing accuracy. Its lead carries over to both external cohorts. On CPTAC and BCNB benchmarks, PairSlide scores 57.64\% and 58.98\%, outperforming the SOTA pathology VLM models.

\section{Related Work}\label{sec:related}
\textbf{Pathology vision--language models and Benchmarks.}
Contrastive pathology encoders, including PLIP \citep{plip}, CONCH \citep{conch}, MUSK \citep{musk}, TITAN \citep{titan}, and PRISM \citep{prism}, provide feature representations for pathology VLMs, which operate either on tiles \citep{llavamed,quiltllava,pathgen,pathchat} or directly on whole slides \citep{slidechat,wsillava,cpathomni}. Recent efforts extend these models with reasoning and reinforcement tuning \citep{pathor1,smartpath,pathvlmr1,scaler1,medvlmr1}, preference optimization \citep{eagle}, agentic control \citep{cpathagent,wsiagents,pathfinder,pathoagentic}, and VLM-augmented MIL \citep{vilamil}; PairSlide retains the full-token LongNet design of SlideChat \citep{longnet,gigapath} and evaluates open-weight representatives across these families. Existing pathology VQA benchmarks cover both tile-level \citep{pathvqa,quiltllava,pathmmu,pmcvqa} and slide-level settings \citep{wsivqa,wsillava,slidechat}, but primarily emphasize accuracy and may overlap with the training data of the evaluated models. In contrast, natural-image benchmarks probe answer priors \citep{vqacp}, hallucination \citep{pope}, adversarial robustness \citep{naturalbench}, or vision-indispensable reasoning \citep{mmstar}, but do not address gigapixel pathology inputs. To our knowledge, CleanSlide is the first slide-level benchmark to jointly evaluate blind performance, blind-screened and prior-resistant subsets, and paired counterfactuals. CAMELYON16 \citep{camelyon} is used only for the motivating study.

\textbf{Do models use the image?} Blind baselines, hallucination probes, and contrast sets demonstrate that vision-language models frequently answer questions without relying on actual visual evidence \citep{pope,chair,vcd}. To mitigate this, preference-based remedies that beyond standard contrastive decoding \citep{vcd}, typically contrast correct and hallucinated responses \citep{povid,rlhfv}, or contrast the original image with a counterfactual baseline. These baselines include randomly cropped versions (mDPO; \citealp{mdpo}), retrieved similar images (Re-Align; \citealp{realign}), nearest real neighbors with symmetric response rankings (SymMPO; \citealp{symmpo}), or minimally edited, largely synthesized counterparts (S-VCO; \citealp{svco}). These counterfactuals are either edited or synthesized,
which is impractical for gigapixel slides whose findings are graded tissue patterns rather than discrete objects, or retrieved by visual similarity, which guarantees neither a changed answer nor matched staining and scanner. Consequently, existing pathology alignment methods optimize responses or rewards directly without utilizing slide-level contrasts \citep{eagle,pathor1,smartpath}. Inspired by \citet{teney2020learning}, who leveraged naturally occurring minimal pairs (such as the complementary images in VQA v2 \citep{vqav2}) as a training signal, Pair-DPO introduces the use of real, naturally occurring pathology pairs: two real slides that share the identical question, cancer type, and tissue source site, but present opposite clinical findings. This approach yields three distinct logical advantages: first, it requires no image editing or synthesis, bypassing the limitations of mDPO-style synthetic counterfactuals, which provide little utility on complex histopathology slides. Second, because both slides originate from the same tissue source site, confounding signatures like site-specific staining protocols and scanner-specific artifacts are shared and cannot be exploited to separate the pair. Third, by holding the question and answer constant while varying only the slide within each paired comparison, any reward component relying solely on the text (such as the clinical prior) mathematically cancels out within each individual training term, ensuring that the clinical prior, what reflecting highly accurate, genuine epidemiology, is neither falsely rewarded nor artificially penalized, unlike in SymMPO where cancellation only occurs in aggregate.

\section{The CleanSlide Benchmark}\label{sec:cleanslide}

\begin{figure}
    \centering
    \includegraphics[width=1\linewidth]{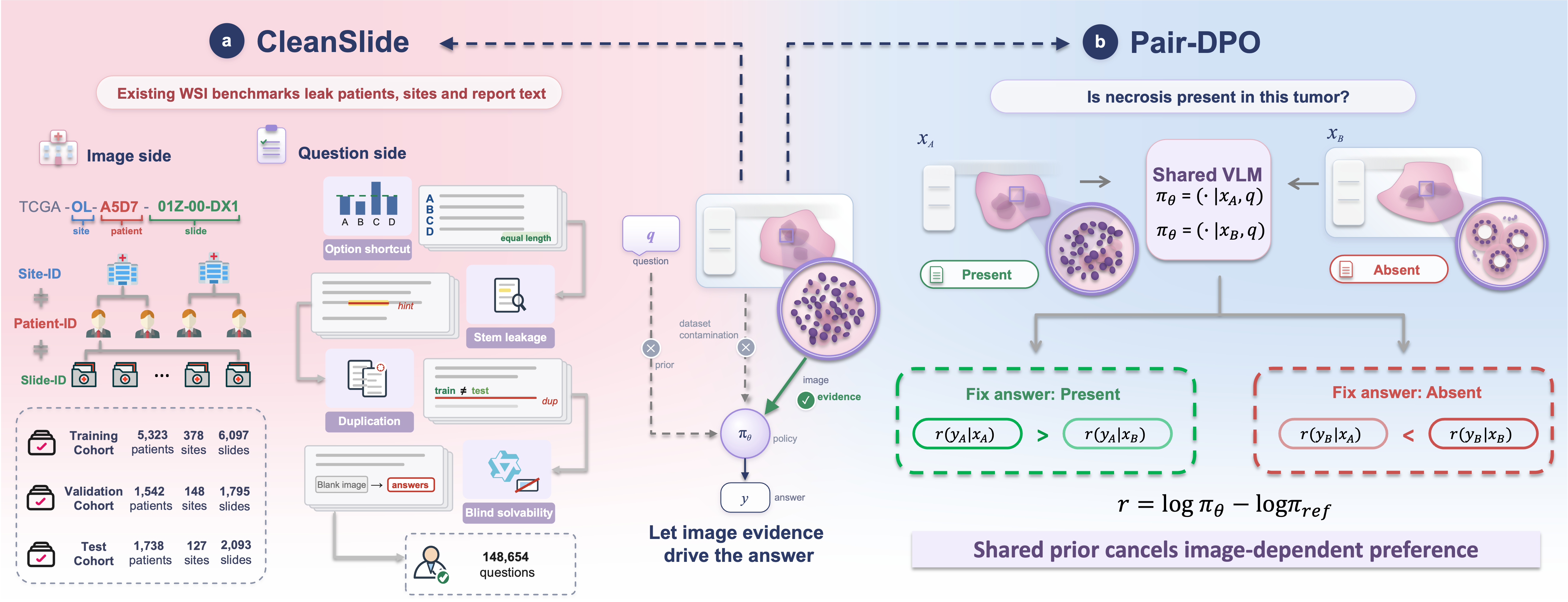}
    \caption{Overview of CleanSlide and Pair-DPO. \textbf{(a)} CleanSlide splits TCGA
slides so that train, validation and test share no patient, site or
slide. Its questions are audited for option shortcuts, stem leakage,
cross-split duplication and blind solvability, and are checked by
pathologists. \textbf{(b)} Pair-DPO asks one question on two real slides of the same
cancer type whose reports give opposite findings. Each answer must be
preferred on its own slide over the other slide, so the prior cancels within
the pair and no image is edited.}
    \label{fig:method}
\end{figure}

\subsection{Benchmark Construction}\label{sec:cs-construction}

As shown in Figure~\ref{fig:method}~(a), CleanSlide contains 148,654 four-option multiple-choice questions over
9,985 diagnostic TCGA slides from 32 cancer types.
We define dataset contamination along two axes: image- and question-side contamination.

On the image side, we split by both patient and tissue source site (TSS).
Previous reseach~\citep{zhang2026auditingdataleakagewholeslide} shows that several public WSI datasets
contain most test cases of existing TCGA-derived benchmarks (e.g., WSI-Bench-
train covers 92.3\% of SlideBench-VQA-TCGA and 100\% of WSI-VQA test cases),
and that TSS identity is linearly decodable from CONCH slide features. Our
patient- and TSS-disjoint splits therefore place models trained on CleanSlide
in the cross-site regime. On the question side, we assess four sources of contamination: option shortcuts, stem leakage, cross-split duplication, and blind solvability, measuring, respectively, answer cues from option structure, answer-revealing stem cues, repeated questions across splits, and whether a text-only LLM can answer above chance without the slide image.
Our questions are collected through two tiers. The first incorporates questions from prior public TCGA benchmarks, including WSI-Bench, SlideBench, and WSI-VQA. The second consists of newly generated questions, with a small templated subset derived from GDC structured labels and a larger set mined from free-text pathology reports using Qwen3-32B~\citep{qwen3}. All questions undergo the same four-dimensional screening. CleanSlide is constructed exclusively from TCGA to avoid cohort leakage and source-specific shortcuts, enabling a cleaner assessment of tissue-grounded transfer while reserving other commonly used data sources, such as CPTAC and BCNB, as external cohorts for evaluating robustness to acquisition shifts. Certified pathologists independently review our benchmark (Appendix~\ref{app:pathologist}).

After curation, CleanSlide achieves a 0\% cross-split duplication rate, with stem leakage and option-shortcut rates of 8.5\% and 25.0\%, respectively. Qwen3-32B achieves 41\% accuracy in the text-only setting, quantifying the clinical prior captured by the questions. We do not treat this prior as contamination to be eliminated: clinical diagnosis is inherently Bayesian, as pathologists combine pre-test probabilities from tumor type and grade with visual evidence~\citep{bours2021bayes}. Thus, question-derived clinical knowledge is legitimate rather than a shortcut~\citep{niu2021counterfactual,teney2020value}. Our focus is instead on the incremental contribution of image evidence, which we isolate by measuring the blind prior and canceling it within paired questions.

\subsection{Metrics}\label{sec:cs-metrics}

We evaluate models at three
increasingly strict layers of prior control, then quantify image reliance
with evidence use and paired counterfactual metrics. \textbf{Layer-A} contains the full test set (28,923 questions) and corresponds
to the conventional benchmark accuracy. \textbf{Layer-B} removes questions solvable from general knowledge. We
screen all questions with three general-purpose VLMs
(Qwen3-VL-8B/32B~\citep{qwen3vl} and LLaVA-OneVision-1.5-8B
\citep{onevision}) using the question alone, and retain the 13,717
questions (47\%) for which their plurality vote is incorrect. \textbf{Layer-C} further removes questions solvable from in-domain priors.
We use a fixed, external blind ensemble covering three sources of prior:
three text-only LLMs (Qwen3-32B~\citep{qwen3}, Qwen2.5-7B-Instruct
\citep{qwen25llm}, and DeepSeek-R1-Distill-7B~\citep{deepseekr1}), two
general-purpose VLMs, and two pathology VLMs
(Qwen3-VL-32B~\citep{qwen3vl}, LLaVA-OneVision-1.5-8B~\citep{onevision},
Patho-R1~\citep{pathor1}, and SlideChat~\citep{slidechat}). Layer-C retains questions
that every blind model answers incorrectly, yielding 2,501 questions shared by all models.

As accuracy alone cannot tell whether a model has read the slide, we report
three metrics of evidence use. The first is Evidence-Weighted Accuracy
(EWA), a new metric we introduce here. Let $\mathrm{Acc}_{\text{image}}$ and
$\mathrm{Acc}_{\text{blind}}$ be the Layer-B accuracy with the image and
with a blank input; the latter, the blind
score, estimates what the model solves from priors alone. Previous
studies report the raw gain $\mathrm{Acc}_{\text{image}}-\mathrm{Acc}_{\text{blind}}$
\citep{mmstar}, which ignores the headroom left by the prior; normalizing
by that headroom instead rewards models that stay inaccurate. EWA weights
the normalized gain by accuracy, so it is high only when a model is both
accurate and uses the slide:
$$\mathrm{EWA} = \mathrm{Acc}_{\text{image}} \times \max\left(0, \frac{\mathrm{Acc}_{\text{image}} - \mathrm{Acc}_{\text{blind}}}{1 - \mathrm{Acc}_{\text{blind}}}\right).$$

The remaining metrics are based on the paired counterfactual test. We construct counterfactual pairs $(x_A, x_B)$ from the pathology reports. The two slides in each pair come from different patients with the same cancer type and the same tissue source site. According to the report, a target finding is present in $x_A$ and absent in $x_B$, so the ground-truth answers $y_A$ and $y_B$ differ. Let $\hat{y}(x,q)$ denote the predicted answer, $p_\theta(\cdot \mid x, q)$ the predicted distribution over the answer options, and $\mathbb{E}$ the mean over pairs. Following~\citep{mdpo}, the paired sensitivity measures the amount of change as the $L_1$ distance between the predicted distributions for the two slides:
\[
\mathrm{Sens} = \mathbb{E}\big[\,\lVert p_\theta(\cdot \mid x_A, q) - p_\theta(\cdot \mid x_B, q)\rVert_1\,\big].
\]
However, a model that reacts to staining differences or noise can also have a large $\mathrm{Sens}$. Hence, a large $\mathrm{Sens}$ alone does not show that the prediction changes because of the target finding. The directional accuracy measures the direction of change:
\[
\mathrm{Dir} = \Pr\big[\hat{y}(x_A,q)=y_A \;\big|\; \hat{y}(x_A,q)\neq\hat{y}(x_B,q),\ y_A\in\{\hat{y}(x_A,q),\hat{y}(x_B,q)\}\big].
\]
This metric considers only the pairs on which the model gives different answers to the two slides and one of the two answers is $y_A$. Among these pairs, $\mathrm{Dir}$ is the fraction on which the model assigns $y_A$ to $x_A$ rather than to $x_B$. If the answer changes are unrelated to the target finding, $y_A$ is equally likely to be assigned to either slide, and $\mathrm{Dir}=50\%$. A model that correctly recognizes the finding approaches $\mathrm{Dir}=100\%$.

Notably, even a model that reads every slide correctly cannot be expected to change its answer on every pair, because pathology-report labels describe the whole specimen while each slide shows only one tissue section. We therefore compare $\mathrm{Sens}$ with control models trained on the same data for the same number of steps. For $\mathrm{Dir}$, the chance reference is $50\%$. Values above $50\%$ indicate that answer changes follow the target finding more often than chance. 

\section{Model and Training}\label{sec:method}
\subsection{Preliminaries}\label{sec:prelim}
A vision--language model with parameters $\theta$ defines a policy $\pi_\theta(y \mid x, q)$ over
answers. Direct preference optimization (DPO) \citep{dpo} aligns such a
policy with pairwise preferences without fitting a separate reward model.
It starts from the KL-regularized objective of reinforcement learning from
human feedback and shows that the optimal policy defines an implicit reward
through its own likelihood,
\begin{equation}
r_\theta(y \mid x, q) \;=\; \log \pi_\theta(y \mid x, q) \;-\;
\log \pi_{\mathrm{ref}}(y \mid x, q),
\end{equation}
where $\pi_{\mathrm{ref}}$ is a frozen reference policy. Given a preferred
answer $y_w$ and a rejected answer $y_l$ for the same input, a
Bradley--Terry model of the preference then yields the DPO loss
\begin{equation}
\mathcal{L}_{\mathrm{DPO}} = \sigma^{+}\!\big(r_\theta(y_w \mid x, q) -
r_\theta(y_l \mid x, q)\big), \qquad \sigma^{+}(u) = \log\big(1 + e^{-\beta u}\big),
\end{equation}
where the temperature $\beta$ sets how sharply the policy may depart from
the reference. 
Since DPO depends only on the relative likelihood of the two
answers, shifting both likelihoods equally does not change the loss. This
removes only what the two answers share: scored on the same slide, they are
still separated by the question's prior, so the loss can fall without
reading the tissue. Response-side variants \citep{povid,rlhfv} share this
weakness. Image-side variants instead edit or synthesize the
rejected image \citep{mdpo,svco}, which gigapixel slides do not permit
faithfully, or retrieve a similar one \citep{realign} whose answer need not
differ. Our Pair-DPO offers a second real slide with the opposite verified
finding.

\subsection{Pair-DPO and PairSlide}\label{sec:pairdpo}

Different from DPO and its variants, Pair-DPO compares two real slides for one
answer. The two slides of a pair share the question and its options, the cancer
type and the tissue site, so none of these can tell the slides
apart. Let $c$ denote the shared context and split the reward into a part
that depends on the slide only through $c$ and a remainder $\delta_\theta$,
\begin{equation}
r_\theta(y \mid x, q) \;=\; \bar{r}_\theta(y, c) \;+\; \delta_\theta(y, x, c),
\label{eq:reward-split}
\end{equation}
where $\bar{r}_\theta$ collects everything the model has learned to say
about $y$ without reading the tissue, such as the prior of the question,
the base rate of the cancer type and the staining style of one
tissue source site. Because $c$ is identical for $x_A$ and $x_B$, each term of
Pair-DPO reduces to
\begin{equation}
r_\theta(y_A \mid x_A, q) - r_\theta(y_A \mid x_B, q) \;=\;
\delta_\theta(y_A, x_A, c) - \delta_\theta(y_A, x_B, c),
\label{eq:prior-cancel}
\end{equation}
and this holds for any such split. A parameter change that shifts the
reward by a function of $c$ alone leaves the difference unchanged, so the
loss cannot be lowered by any signal that the two slides share, whether it
comes from the text or from the image. Its gradient falls entirely on the
tissue that differs between them. A response-side comparison of two
answers on one slide instead keeps $\bar{r}_\theta(y_w, c) -
\bar{r}_\theta(y_l, c)$ and can be lowered by learning the prior.

As shown in Figure \ref{fig:method}~(b), we use the same
reward of Section~\ref{sec:prelim} and take as the reference the same network with the adapter disabled, so no second model has to be stored.
For a pair $(x_A, x_B)$ with opposite golds $(y_A, y_B)$ we ask that $y_A$
be better supported on the slide where it is true than on the slide where
it is false, and symmetrically for $y_B$,
\begin{equation}
\mathcal{L}_{\mathrm{pair}} = \tfrac{1}{2}\,\sigma^{+}\!\big(
r_\theta(y_A \mid x_A) - r_\theta(y_A \mid x_B)\big)
+ \tfrac{1}{2}\,\sigma^{+}\!\big(
r_\theta(y_B \mid x_B) - r_\theta(y_B \mid x_A)\big).
\end{equation}
The loss is small when swapping the slide for its counterfactual moves the
model's preference in the direction of the evidence. Each term holds the
answer fixed and varies the input, as in mDPO, but both inputs are real
slides with opposite verified findings. The two slides also share the
question and the cancer type. An answer that ignores the slide, therefore
raises both rewards equally in a term and cannot lower the loss.

Based on the proposed Pair-DPO, we design and train PairSlide. Inspired by the design
of SlideChat \citep{slidechat}, PairSlide adopts a frozen CONCH
encoder \citep{conch} that produces one 512-dimensional feature per tile, and a
two-layer LongNet \citep{longnet} with 4.2M parameters, which contextualizes the
tile sequence. It also includes a two-layer projector with 14.7M parameters, which maps each token
into the language-model space. All tile tokens are consumed by
Qwen2.5-7B-Instruct \citep{qwen25llm} through a rank-16 LoRA adapter
\citep{lora} with 40.4M parameters, so 59.3M parameters train in total. The full training procedure is shown in Appendix~\ref{app:procedure}.

\section{Experiments}\label{sec:exp}
\subsection{Experimental setup}

We evaluate on the held-out CleanSlide test split at all three layers, and test transfer beyond TCGA on two external
cohorts, BCNB and CPTAC, using their SlideBench-VQA question sets.
Every confidence interval
is a 95\% bootstrap interval clustered by patient. Following~\cite{slidechat}, tile-level models that consume raw images are evaluated with 30 tiles and a majority vote, and every model is scored under one
constrained protocol, in which the predicted answer is the option letter
with the highest logit among the options the question offers. Our
models were trained on two H200 GPUs for 44.5 hours, with a learning rate of $2{\times}10^{-5}$, batch
composition with $\beta{=}1$, $\lambda{=}5$, and a
pair rate of $\rho{=}0.26$ per step. Appendix~\ref{app:procedure} shows the full parameter settings.

\subsection{Published models earn their accuracy from the prior}
\label{sec:exp-main}

\begin{table}[t]\centering\small
\caption{Comparison of general-purpose VLMs, medical and pathology VLMs,
and PairSlide on the CleanSlide test set. SlideChat$^{*}$ is a
full-parameter control trained on the CleanSlide training split. Patient-clustered 95\% CIs are shown in brackets. In every
column except Blind, \textbf{bold} marks the best and \underline{underline}
the second-best model.}
\label{tab:main}
\begin{adjustbox}{max width=\linewidth}
\begin{tabular}{lcccccc}
\toprule
Model & Proto & Layer-A & Layer-B & Layer-C & Blind & EWA \\
\midrule
Qwen3-VL-32B \citep{qwen3vl}
& tile & 55.23 [54.50--55.96] & 26.61 [25.71--27.49]
& 11.49 [10.23--12.93] & 16.66 & 3.18 \\

Qwen3-VL-8B \citep{qwen3vl}
& tile & 48.36 [47.62--49.10] & 22.06 [21.16--22.95]
& 11.79 [10.47--13.12] & 3.06 & 4.32 \\

LLaVA-OneVision-1.5-8B \citep{onevision}
& tile & 51.95 [51.28--52.62] & 25.87 [25.02--26.74]
& 6.52 [5.59--7.55] & 16.13 & 3.00 \\

Qwen2.5-VL-7B \citep{qwen25}
& tile & 46.31 [45.61--47.01] & 22.47 [21.66--23.26]
& 9.98 [8.82--11.22] & 18.90 & 0.99 \\

LLaVA-Med \citep{llavamed}
& tile & 32.53 [31.94--33.13] & 32.92 [32.11--33.72]
& 21.36 [19.66--23.05] & 32.90 & 0.01 \\

PathGen-LLaVA \citep{pathgen}
& tile & 49.23 [48.55--49.93] & 39.53 [38.64--40.45]
& 18.71 [17.09--20.28] & 36.83 & 1.69 \\

Quilt-LLaVA \citep{quiltllava}
& tile & 41.40 [40.74--42.05] & 31.93 [30.99--32.86]
& 20.00 [18.45--21.65] & 31.54 & 0.18 \\

Patho-R1 \citep{pathor1}
& tile & 46.71 [45.96--47.42] & 41.67 [40.68--42.67]
& 16.03 [14.57--17.52] & 39.75 & 1.33 \\

ScaleReasoner-R1 \citep{scaler1}
& tile & 27.67 [27.10--28.19] & 21.80 [21.08--22.51]
& 13.27 [11.93--14.64] & 21.70 & 0.03 \\

MedVLM-R1 \citep{medvlmr1}
& tile & 38.23 [37.57--38.89] & 27.39 [26.54--28.24]
& 11.68 [10.40--12.96] & 27.94 & 0.00 \\

WSI-LLaVA \citep{wsillava}
& slide & 54.00 [53.29--54.76] & 46.62 [45.69--47.55]
& 30.76 [28.88--32.58] & 43.44 & 2.62 \\

SlideChat \citep{slidechat}
& slide & 62.03 [61.31--62.70] & 44.72 [43.78--45.67]
& 8.76 [7.63--9.91] & 41.01 & 2.81 \\
\midrule
SlideChat$^{*}$ (full-parameter control, 7.65B)
& slide & \underline{75.18 [74.51--75.82]} & \underline{62.70 [61.69--63.66]}
& \underline{42.86 [40.85--45.08]} & 56.12 & \underline{9.40} \\

\textbf{PairSlide}
& slide & \textbf{77.79 [77.14--78.40]} & \textbf{67.35 [66.43--68.26]}
& \textbf{46.78 [44.70--49.22]} & 57.76 & \textbf{15.29} \\

\bottomrule
\end{tabular}
\end{adjustbox}
\end{table}

Table~\ref{tab:main} reports accuracy at three layers together with the
blind score and EWA, so that a model's prior knowledge and its use of image
evidence can be read side by side. Three observations follow. First,
PairSlide attains the highest accuracy at all three layers, 77.79\%,
67.35\%, and 46.78\%, together with the highest EWA, 15.29. No published
model exceeds an EWA of 4.32, and the strongest slide-level models answer
almost as well without the slide as with it: WSI-LLaVA and SlideChat reach
blind scores of 43.44\% and 41.01\% against Layer-B accuracies of 46.62\%
and 44.72\%, for EWA of 2.62 and 2.81. PairSlide has both the highest blind
score, 57.76\%, and the highest EWA, so its accuracy is not explained by
prior knowledge alone.
Second, reinforcement learning does not by itself produce use of the image
under our protocol. ScaleReasoner-R1, Patho-R1, and MedVLM-R1 reach EWA of
only 0.03, 1.33, and 0.00, because their blind scores (21.70\%, 39.75\%,
27.94\%) nearly match or exceed their Layer-B accuracy (21.80\%, 41.67\%,
27.39\%), which is therefore almost entirely prior knowledge. A reward
defined on the final answer can be satisfied by sharpening that prior,
which may explain why reasoning-oriented post-training improves the answer
without improving slide use.
Third, we add SlideChat$^{*}$, which updates all its parameters on the
CleanSlide split under the released SlideChat schedule and so shares our
training data with a larger trainable set. It reaches 75.18\%, 62.70\%, and
42.86\% at the three layers with an EWA of 9.40, all below PairSlide. With
the training data held fixed, the remaining difference reflects the
training recipe and is widest in EWA, 15.29 against 9.40.

\subsection{Pair-DPO improves evidence use}
\label{sec:exp-ablation}

\begin{table}[t]\centering\footnotesize\setlength{\tabcolsep}{4pt}
\caption{Step-matched ablation. Every row is a run of 2,000
steps from the same SFT base. Brackets are patient-clustered
95\% CIs.}
\label{tab:ablation}
\begin{adjustbox}{max width=\linewidth}
\begin{tabular}{lcccccc}
\toprule
 & \multicolumn{3}{c}{Loss terms} & & & \\
\cmidrule(lr){2-4}
Method & CE & mDPO pair & real pair & EWA & Dir & Sens \\
\midrule
\multicolumn{7}{l}{\textit{Step Fixed.}} \\
SFT base & -- & -- & -- & 12.70 [9.91--15.49] & 59.1 [51.9--66.1] & 0.072 [0.067--0.080] \\
CE & \checkmark & -- & -- & 12.55 [9.70--14.80] & 60.5 [55.4--66.4] & 0.084 [0.080--0.088] \\
mDPO & \checkmark & \checkmark & -- & 13.94 [10.98--16.91] & 60.7 [54.5--67.0] & 0.090 [0.085--0.096] \\
\textbf{Pair-DPO} & \checkmark & -- & \checkmark & \textbf{15.29 [12.38--18.21]} & \textbf{65.5 [60.8--70.2]} & \textbf{0.108 [0.102--0.114]} \\
\midrule
\multicolumn{7}{l}{\textit{Pairs only.}} \\
mDPO & -- & \checkmark & -- & 13.88 [11.13--16.63] & 57.7 [50.0--65.1] & 0.067 [0.062--0.073] \\
\textbf{Pair-DPO} & -- & -- & \checkmark & \textbf{16.69 [13.96--19.43]} & \textbf{64.8 [61.2--68.4]} & \textbf{0.190 [0.180--0.202]} \\
\bottomrule
\end{tabular}
\end{adjustbox}
\end{table}

Table~\ref{tab:ablation} isolates the effect of pair supervision. All trained runs start from the same SFT checkpoint and train for 2,000 steps. We first compare runs with a matched pair exposure of 26\%. The Pair-DPO raises $\mathrm{EWA}$ from 12.55 to 15.29.
Since both runs use the same number of steps, the gain comes from pair
supervision rather than from additional optimization. In
contrast, mDPO reaches only 13.94. mDPO and Pair-DPO see pairs equally often, so pair
exposure cannot explain the difference between them. To rule out that mDPO simply has not
converged at this pair rate, the pairs-only runs spend all 2,000 steps on
the pair loss, giving it about four times as many pair updates.
Pair-DPO uses real
pairs and reaches 0.190, the highest $\mathrm{Sens}$ in the
table, together with the highest $\mathrm{EWA}$ of 16.69. mDPO falls from
0.090 to 0.067. These results show that counterfactual pairs do not help
uniformly and that their effect depends on how the pairs are constructed.
mDPO builds a counterfactual sample by removing pathology-related tiles. This
removal introduces visual differences that are unrelated to the
target finding. The model can then separate the two images by these
artifacts instead of comparing the evidence, and training on mDPO pairs
alone appears to strengthen this shortcut. Synthetic counterfactual samples may
therefore provide not only weaker supervision but also a misleading signal
that shifts learning away from evidence-based reasoning.

\begin{figure}
    \centering
    \includegraphics[width=1\linewidth]{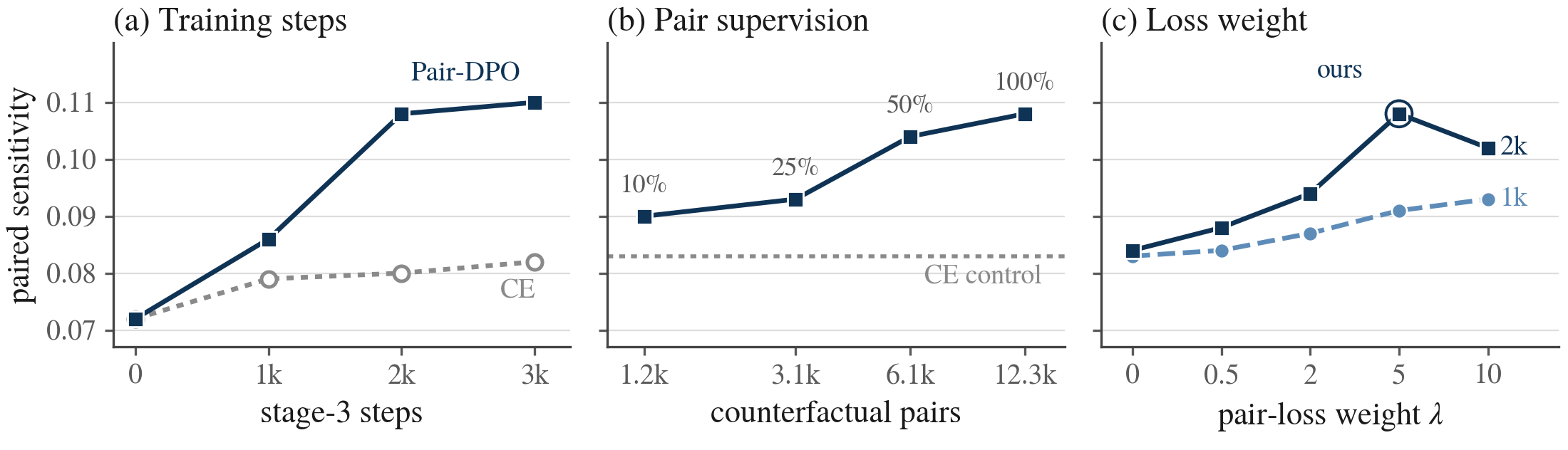}
    \caption {Ablations. \textbf{(a)} Sensitivity against the number of steps for the cross-entropy control and for Pair-DPO.  \textbf{(b)} Sensitivity against the number of real counterfactual pairs available for training. \textbf{(c)} Sensitivity against the pair-loss weight $\lambda$ at two step budgets.}
    \label{fig:ablation}
\end{figure}

Figure~\ref{fig:ablation} shows Pair-DPO's sensitivity to training budget, pair supervision, and $\lambda$. First, training steps alone have limited effect: cross-entropy training increases sensitivity only from 0.072 to 0.082 over 3{,}000 updates, whereas Pair-DPO shows a larger gain that largely plateaus after 2{,}000 updates. We therefore use 2{,}000 updates. Second, sensitivity increases consistently with the amount of pair supervision; even 10\% of the pairs outperforms the cross-entropy control, and the full pool does not saturate, indicating that pair supervision remains a limiting factor. Third, the effect of training budget depends on $\lambda$: the two budgets are nearly identical at $\lambda=0$, diverge as $\lambda$ increases, and show the largest gap around $\lambda=5$. Beyond $\lambda=5$, additional updates provide little benefit. We thus use $\lambda=5$ and 2{,}000 updates.

\subsection{Evidence use on the external cohorts}
\label{sec:exp-external}

\begin{table}[t]\centering\small
\caption{Whole-slide VQA on the SB-BCNB (SlideBench-VQA-BCNB) full set of
7,274 questions and the external SB-CPTAC (SlideBench-VQA-CPTAC) cohort of
240 questions. SlideChat is our rerun of the public checkpoint under the
matched protocol. \textbf{Bold} marks the
best and \underline{underline} the second-best model.}
\label{tab:wsivqa}
\begin{adjustbox}{max width=\linewidth}
\begin{tabular}{lccccccc}
\toprule
& & \multicolumn{3}{c}{SB-BCNB} & \multicolumn{3}{c}{SB-CPTAC} \\
\cmidrule(lr){3-5}\cmidrule(lr){6-8}
Model & Proto & Acc & Blind & EWA & Acc & Blind & EWA \\
\midrule
Quilt-LLaVA
& tile & 42.52 [41.50--43.57] & 43.73 [42.70--44.77] & 0.00
& \underline{47.92 [41.66--54.17]} & 44.58 [38.33--50.83] & 2.89 \\

LLaVA-Med
& tile & 35.19 [34.11--36.28] & 35.08 [33.98--36.15] & 0.06
& 26.67 [21.25--32.50] & 25.83 [20.42--31.67] & 0.30 \\

PathGen-LLaVA
& tile & 44.97 [43.75--46.01] & 45.92 [44.80--46.94] & 0.00
& 31.25 [25.42--37.09] & 26.67 [21.25--32.08] & 1.95 \\

Patho-R1
& tile & 48.21 [47.05--49.31] & 49.41 [48.33--50.54] & 0.00
& 21.25 [16.25--26.67] & 16.25 [12.08--21.26] & 1.27 \\

EAGLE
& tile & \underline{55.64 [54.54--56.85]} & 42.81 [41.82--43.84] & \textbf{12.48}
& 44.17 [37.50--50.83] & 17.50 [12.91--22.08] & \underline{14.28} \\

SmartPath-R1
& tile & 48.16 [47.19--49.14] & 51.21 [50.21--52.21] & 0.00
& 29.58 [24.17--35.42] & 35.42 [29.17--42.08] & 0.00 \\

ScaleReasoner-R1
& tile & 49.57 [48.47--50.71] & 49.45 [48.38--50.49] & 0.12
& 33.33 [27.08--39.58] & 22.50 [17.50--27.92] & 4.66 \\

MedVLM-R1
& tile & 32.55 [31.79--33.31] & 27.74 [26.85--28.72] & 2.17
& 13.33 [9.17--17.51] & 15.00 [10.42--19.59] & 0.00 \\

WSI-LLaVA
& slide & 46.27 [45.09--47.39] & 44.58 [43.47--45.67] & 1.41
& 43.75 [37.08--50.00] & 35.83 [30.00--42.08] & 5.40 \\

SlideChat
& slide & 50.91 [50.02--51.81] & 44.76 [43.69--45.81] & 5.67
& 42.50 [36.25--48.76] & 37.08 [30.83--43.33] & 3.66 \\
\midrule
\textbf{PairSlide}
& slide & \textbf{58.98 [57.88--59.99]} & 49.41 [48.38--50.41] & \underline{11.16}
& \textbf{57.64 [51.80--63.33]} & 42.36 [36.53--48.19] & \textbf{15.28} \\

\bottomrule
\end{tabular}
\end{adjustbox}
\end{table}

We evaluate out of domain on CPTAC and BCNB, using the official
SlideBench-VQA-CPTAC and
SlideBench-VQA-BCNB set with benchmark-provided features
(Table~\ref{tab:wsivqa}). On SB-BCNB, PairSlide achieves the highest
accuracy (58.98\%) and exceeds the confidence interval of every baseline.
Access to the slide does not guarantee its use: four of the eight
tile-level models score lower with the slide than without it, so their EWA
is zero, and SmartPath-R1 drops from 51.21\% to 48.16\%. EAGLE shows the
opposite profile, with the highest EWA on SB-BCNB (12.48 vs.\ 11.16 for
PairSlide) but lower accuracy (55.64\% vs.\ 58.98\%), since it starts from a
lower blind accuracy (42.81\% vs.\ 49.41\%). On SB-CPTAC, PairSlide reaches
57.64\% [51.80--63.33] and
the highest EWA (15.28, ahead of EAGLE at 14.28). The smaller sample
($n{=}240$) widens the intervals, and PairSlide is distinguishable from all
baselines except Quilt-LLaVA, whose accuracy comes mostly from the prior
(blind 44.58\%, EWA 2.89). PairSlide is the only model in the top two for
both accuracy and EWA on both cohorts, suggesting that prior knowledge and
image evidence are complementary.

\subsection{Does pair-DPO transfer beyond pathology?}
\label{sec:exp-transfer}
\begin{figure}
    \centering
    \includegraphics[width=1\linewidth]{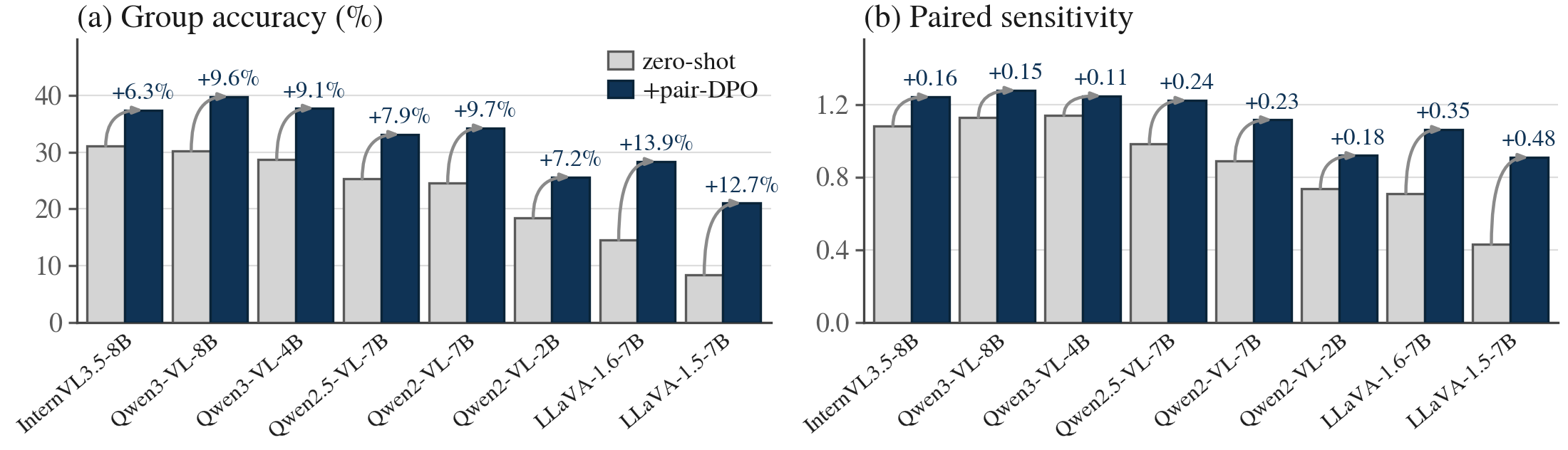}
    \caption{Natural-image transfer on the decontaminated NaturalBench test (1,182 pairs, 585 complete groups). G-acc is the official group accuracy and Sens the paired sensitivity.}
    \label{fig:tranfer}
\end{figure}

We finally ask whether the loss is specific to pathology, using NaturalBench
\citep{naturalbench}, whose items pair one question with two images whose
correct answers differ. We apply the same loss and $\lambda$ to eight vision--language models from five families
\citep{qwen3vl,internvl,qwen25,qwen2vl,llava15,llavanext}, training on
NaturalBench pairs and testing on the 1,182 pairs whose images never appear
in training. Baselines are a step-matched cross-entropy control and three
step-matched alternatives: a retrieved negative image \citep{realign},
symmetric response ranking \citep{symmpo} and blind-anchored contrast
\citep{svco} (Appendix~\ref{app:transfer} shows the full results). Over zero-shot, pair-DPO raises
group accuracy by 6.33 to 13.85 points and sensitivity by 0.107 to 0.478
(Figure~\ref{fig:tranfer}), most on the two LLaVA bases. Over CE control, Pair-DPO raises sensitivity by 6\% to 25\%,
with non-overlapping intervals on six of eight bases, and per-question
accuracy by 0.17 to 1.65 points. Among all objectives it has the highest
sensitivity on every base and the best group accuracy on five,
while the retrieved negative, whose answer need not differ, is generally the
weakest. The sensitivity gain thus transfers beyond pathology.

\section{Conclusion}

Evaluating whole-slide VLMs is confounded by dataset contamination and reliance on clinical priors. We introduce CleanSlide, a benchmark that disentangles these factors via contamination-controlled splits and question-level audits. Our experiments show that existing models largely fail on prior-resistant subsets, whereas our proposed method yields stronger visual grounding, achieving a 15.29\% visual evidence gain compared with 2.81\% for the strongest baseline. Pair-DPO also improves counterfactual sensitivity and image evidence use ability without compromising accuracy. These gains transfer to external clinical cohorts (SB-BCNB, SB-CPTAC) and natural images (NaturalBench), consistently improving sensitivity across eight general-purpose VLMs. Ultimately, our findings demonstrate that natural counterfactual pairs offer a more reliable and robust supervision signal for evidence-based multimodal reasoning than synthetic or random alternatives, though addressing the remaining limits of counterfactual responsiveness, expanding external validation cohorts, and mitigating the computational cost of in-domain training remain important avenues for future work. 

\bibliography{refs}
\bibliographystyle{plainnat}

\newpage
\appendix

\section{CleanSlide: Composition, Audits and Comparison with Existing Benchmarks}
\label{app:cleanslide}

\begin{figure}[t]
    \centering
    \includegraphics[width=1\linewidth]{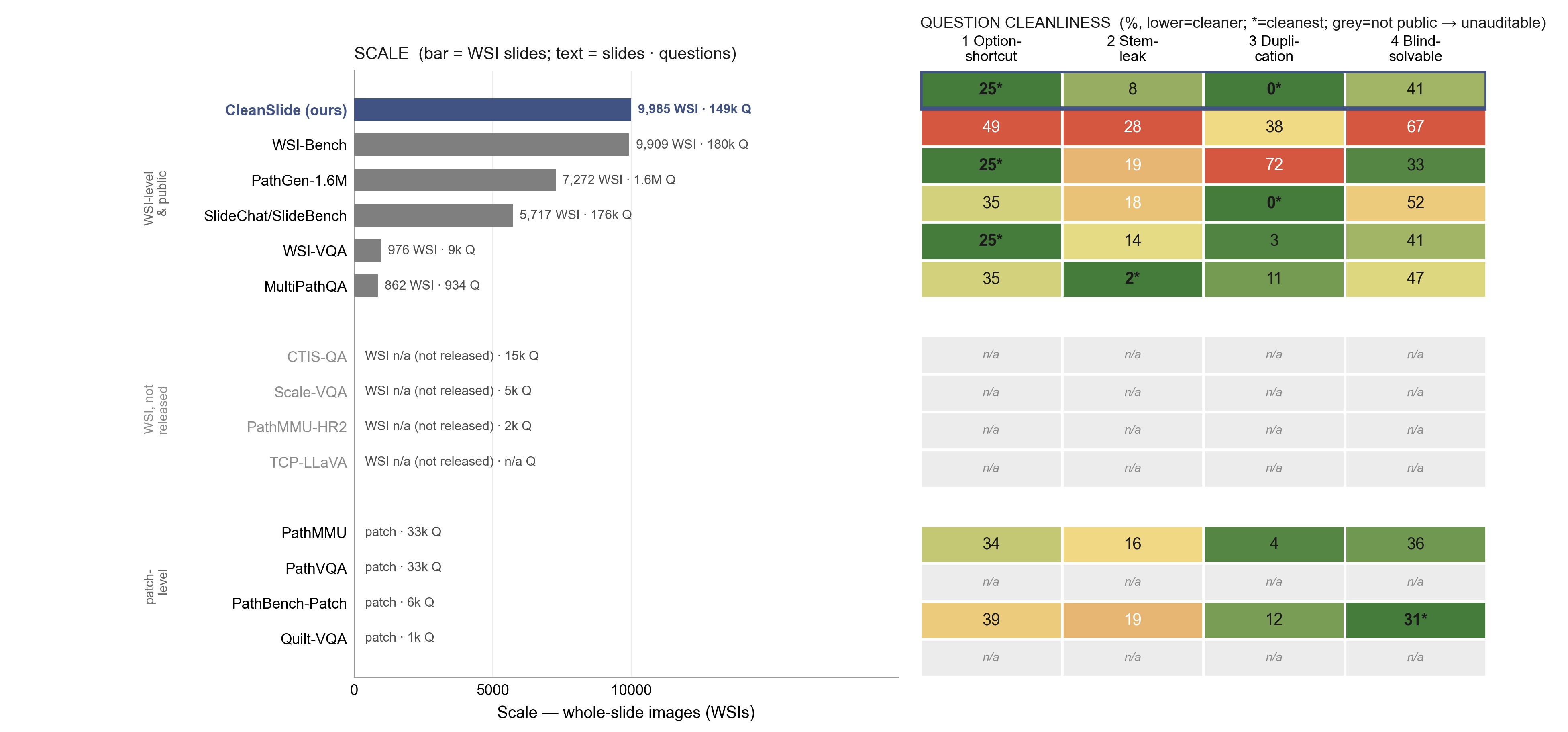}
    \caption{Cross-benchmark comparison. \textbf{Left:} scale of pathology VQA
    benchmarks; the bar is the number of whole-slide images and the text gives
    slides and questions. \textbf{Right:} question cleanliness under the four
    audits of Section~\ref{sec:cs-construction}, in \%, lower is cleaner; $*$
    marks the cleanest value in each column. Blind solvability is the text-only
    accuracy of Qwen3-32B. Grey cells are benchmarks that cannot be audited
    because their data are not released.}
    \label{fig:crossbench}
\end{figure}

CleanSlide uses 9,985 diagnostic TCGA slides, each with its pathology report and
its patient and tissue-source-site identifiers. The tissue source site is the
institution that submitted the specimen, so splitting on it removes the staining
and scanner signatures by which a model could recognise a split. Train, validation
and test share no slide, patient or tissue source site;  the counterfactual pairs of
Section~\ref{sec:cs-metrics} are built inside each split, so they are
patient-disjoint by construction. Questions come from two tiers. The first reuses
the TCGA questions of WSI-Bench \citep{wsillava}, SlideBench \citep{slidechat} and
WSI-VQA \citep{wsivqa}, placed in the CleanSlide splits according to their slides
and audited like any other question, which removes the cross-split duplicates that
the original releases contain. The second is new: a small templated subset derived
from structured GDC labels, and a larger subset mined from the free-text pathology
reports with Qwen3-32B \citep{qwen3}, covering standardised findings such as
lymphovascular invasion, necrosis, perineural invasion, margin status and grade.
All questions are four-option multiple choice.

Each audit is a rate over questions, lower is cleaner, and each is run with the
same protocol on every public benchmark in Figure~\ref{fig:crossbench}. Option
shortcut measures answer cues in the option structure (25.0\% on CleanSlide).
Stem leakage is the fraction of questions whose stem alone reveals the answer
(8.5\%). Cross-split duplication is the fraction of test questions that also occur
in the training split (0.00\%). Blind solvability is the accuracy of text-only
Qwen3-32B given the stem and the options without the slide (41\%). The fourth rate
is measured rather than minimised: the clinical prior a question carries is
legitimate, so we report it for the accuracy to be read against and cancel it
within the counterfactual pairs, rather than deleting the questions a language
model can answer.

Figure~\ref{fig:crossbench} places CleanSlide among fourteen benchmarks. It is the
largest WSI-level set with public slides, 9,985 against 9,909 for WSI-Bench, 7,272
for PathGen-1.6M \citep{pathgen} and 5,717 for SlideBench, and it is the only one
that has the lowest option-shortcut rate and zero duplication at the same time
while keeping stem leakage in single digits. WSI-Bench, the closest in scale, has
49\% option shortcut, 28\% stem leakage, 38\% duplication and a 67\% blind score,
so two-thirds of its questions are answerable without the slide; PathGen-1.6M has
ten times more questions but 72\% of them duplicated across splits; SlideBench
shares the 0\% duplication but has 35\% option shortcut and a 52\% blind score.
MultiPathQA has the lowest stem leakage (2\%) but under a thousand questions, and
PathBench-Patch the lowest blind score (31\%) but is patch-level. Four WSI-level
benchmarks, CTIS-QA, Scale-VQA, PathMMU-HR2 and TCP-LLaVA, have not released their
data and cannot be audited at all, which is why CleanSlide is built from a single
fully documented source: provenance is what makes the audits and the disjoint
splits possible. CPTAC and BCNB are held out entirely as external cohorts
(Section~\ref{sec:exp-external}).

\section{Quantifying Benchmark Contamination} \label{app:contamination}

Accuracy on TCGA-derived WSI benchmarks can reflect contamination rather than evidence use, because public instruction-tuning corpora and evaluation benchmarks are drawn from the same TCGA pool. \citep{zhang2026auditingdataleakagewholeslide} quantified this by tracing canonical case and tissue source site (TSS) identifiers across eight public instruction-tuning corpora and four TCGA-derived benchmarks. Patient-level overlap is extreme: WSI-Bench-train covers 92.3\% of the SlideBench-VQA-TCGA cases, 100\% of the WSI-VQA test cases, and 83.1\% of its own test split, and PathGen-1.6M covers 70.8--94.3\% of the cases in each of the four benchmarks. Removing shared patients does not remove the site confound: WSI-VQA-train and WSI-VQA-test share no patients, but every test TSS also occurs in training, and SlideInstruction+, the only corpus with no case overlap against any of the four benchmarks, still shares 10.1--11.3\% of the TSSs of three of them. Site identity is also linearly decodable from frozen slide embeddings, with a macro AUROC of 0.955 for CONCH-v1 and 0.980 for TITAN. These overlaps coincide with the strongest published results: the highest reported accuracies on SlideBench-VQA-TCGA, 82.5\% for WSI-LLaVA and 89.6\% for MLLM-HWSI, come from checkpoints whose training corpora cover 92.3\% of its cases, and WSI-LLaVA scores higher on patient-leaked than on audit-clean questions of the same benchmark.

These findings shape both the construction of CleanSlide and how we read its results. The CleanSlide training, validation, and test splits share no patient and no TSS, so models trained on the CleanSlide training split, namely SlideChat$^{*}$, PairSlide, are tested on unseen patients from unseen sites. Published checkpoints are in a different position. The imported tier of CleanSlide comes from SlideBench, WSI-Bench, and WSI-VQA, the benchmarks audited above, and WSI-LLaVA, SlideChat, and PathGen-LLaVA were trained on WSI-Bench-train, SlideInstruction, and PathGen-1.6M, respectively. Their CleanSlide accuracy may therefore be optimistic, and their blind accuracy may include recall of training records in addition to prior knowledge, so we read both as upper bounds. Because site identity is decodable from embeddings of the CONCH encoder that PairSlide also uses, we draw each counterfactual pair from a single TSS, so that the staining signature of a site cannot separate the two slides. Finally, the external CPTAC and BCNB cohorts share no patient or TSS with any audited corpus \citep{zhang2026auditingdataleakagewholeslide}, which makes them the contamination-free comparison for published checkpoints.
 
\section{How the four question audit dimensions are cleaned}\label{app:audit}
This appendix describes how each question is measured and repaired, and
shows the effect on real questions. All four steps are text-only and no image is shown to any model.

The first dimension is the option shortcut, which asks whether the answer can
be guessed from the shape of the options rather than from the tissue. Two
cues matter. The answer letter is fixed by a deterministic per-question
permutation, so the released letters are almost uniform at 37,154 for A,
37,044 for B, 37,184 for C and 37,272 for D. Option length is the second cue,
and a question is flagged when the correct option is the uniquely longest one.
Every flagged question goes back to Qwen3-32B with its correct option marked
and with the instruction to shorten all options to a common length while
preserving each meaning and keeping the same option correct. This rewrote
25,496 questions. 

The second dimension is stem leakage, which asks whether the question text
already states the finding. A rule-based detector flags four categories,
namely molecular or immunohistochemical terms, stage and grade words such as
moderately differentiated or Gleason, phrases that announce a finding such as
showing or consistent with, and phrases that state the diagnosis. It also
flags a stem that shares two or more content words with the correct option.
Flagged stems are rewritten by the same model, which may only edit the stem
and may not touch the options or the answer, so the gold answer is unchanged
by construction. Of the flagged questions, 12,534 were rewritten.

The third dimension is cross-split duplication. Each question is reduced to a
signature that holds its normalized stem, its sorted normalized options and
its normalized correct option, so a question that reappears with shuffled
options is caught. Duplicate signatures are removed, with priority to the
earlier pool, and the release is re-checked. The duplication rate is 0.00\%,
and the same check confirms that no patient and no tissue source site occurs
in two splits.

The fourth dimension is blind solvability. Qwen3-32B answers every question
without the image, so the labels
are deterministic. A question it answers correctly is solvable from the text
and from world knowledge alone. About 41\% of the pool falls in that group.
These questions are not deleted, because they remain valid clinical
questions, and deleting them would hide the prior rather than measure it.
They are instead used to carve the evaluation layers of
Section~\ref{sec:cs-metrics}, and every accuracy in the paper is reported
next to a blind score.

The example below shows one stem-leak repair and one option-length repair.

{\footnotesize
\begin{verbatim}
Stem leakage, flagged on "moderate differentiation"
  before: How would you classify the grade of a lung tumor observed
          in the given slide showing moderate differentiation and
          signs of keratinization?
  after : How would you classify the grade of the lung tumor
          observed in the given slide?
  options: G1 | G3 | G4 | *G2*   (unchanged)

Option shortcut, flagged because the correct option is the longest
  before: Highly organized with uniform alveolar spaces
        | Normal lung tissue without any notable aberrations
        | *Disorganized tissue architecture with prominent mucin
           pools and irregular acinar formations*
        | Dense and fibrous with minimal alveolar space
  after : Uniform alveolar structure | No visible abnormalities
        | *Mucin pools and irregular acini*
        | Fibrous with few alveoli
\end{verbatim}
}

\section{Training procedure} \label{app:procedure}
Our method follows a three-stage training scheme. Stages one and two follow the public SlideChat recipe, namely caption
alignment followed by supervised fine-tuning on the complete CleanSlide
training split. Stage three runs 2,000 steps of Algorithm~\ref{alg:stage3}.
A step draws a counterfactual pair with probability $\rho{=}0.26$ and
otherwise two ordinary training questions. On a pair step the loss is the
answer cross-entropy of both slides plus $\lambda\,\mathcal{L}_{\mathrm{pair}}$,
and on the other steps it is the cross-entropy alone. The step-matched
control runs the identical loop with $\lambda{=}0$, so the two arms see the
same slides, the same questions and the same number of updates and differ
only in the preference term.

\begin{algorithm}[t]
\caption{Stage-3 pair-DPO. The step-matched CE control uses $\lambda{=}0$.}
\label{alg:stage3}
\begin{algorithmic}[1]
\State \textbf{input} SFT checkpoint $\theta$, pairs $\mathcal{P}$, singles $\mathcal{S}$, weight $\lambda$, pair rate $\rho{=}0.26$, temperature $\beta{=}1$
\For{step $= 1 \dots 2000$}
  \If{$u \sim \mathrm{U}(0,1) < \rho$}
    \State sample $(x_A, x_B, y_A, y_B, q) \sim \mathcal{P}$
    \State $\mathcal{L} \gets \mathcal{L}_{\mathrm{CE}}(x_A, y_A) / 2 + \mathcal{L}_{\mathrm{CE}}(x_B, y_B) / 2 + \lambda\, \mathcal{L}_{\mathrm{pair}}$
  \Else
    \State sample two singles $(x, y, q) \sim \mathcal{S}$
    \State $\mathcal{L} \gets$ mean $\mathcal{L}_{\mathrm{CE}}$ over the two singles
  \EndIf
  \State accumulate gradients. Update (AdamW, lr $2{\times}10^{-5}$)
\EndFor
\end{algorithmic}
\end{algorithm}

We select $\lambda$ on the validation split with a rule fixed in advance.
Among the arms whose validation accuracy is not significantly below the
$\lambda{=}0$ control, the rule takes the one with the largest validation
sensitivity. It selected $\lambda{=}5$ with 2,000 steps. We then
retrained the selected configuration with three seeds.

\section{Pathologist review of the imported questions}\label{app:pathologist}
 We  asked pathologists to review the imported questions against their slides.
This appendix reports the protocol, the quality scores and what the review
implies for the results of the paper. The review covers the imported tier, namely the
released questions that come from WSI-Bench, SlideBench and WSI-VQA. The
reviewed questions form a stratified sample that is balanced across the
seven task families of this tier and spans the training, validation and
test splits. Each question was shown in a web form with its stem, its four
options and the barcode of its diagnostic slide, which opens in the GDC
slide viewer. The marked answer stayed hidden until the reviewer chose to
reveal it. Reviewers first judged whether the question could be answered
from the text alone and only then opened the slide. Every question was
scored on the seven criteria of Table~\ref{tab:patho-codebook}.

Expert review largely validates the quality of the imported questions. Reviewers found no cases of answer leakage or broken wording, and most questions had plausible distractors and clinically meaningful, unambiguous formulations. The main weaknesses are concentrated in a small set of staging, molecular-subtype, and treatment questions, whose answers depend on clinical information not observable from the tissue image alone. These questions constitute only a small fraction of the released and test sets. Importantly, excluding or rescoring them changes model accuracy only marginally, while such image-unanswerable questions cannot provide evidence of tissue-based reasoning and instead primarily test prior knowledge. Our blind-score evaluation explicitly accounts for this issue, and the paired counterfactual analysis used to assess evidence use relies on tissue-derived findings and is largely unaffected by these flagged families. Overall, the review supports the validity of our main conclusions.

\begin{table}[t]\centering\footnotesize\setlength{\tabcolsep}{4pt}
\caption{Review criteria and codes. A clean question meets the target code
on C1 to C6, and INC records the reviewer's overall decision.}
\label{tab:patho-codebook}
\begin{adjustbox}{max width=\linewidth}\begin{tabular}{llll}
\toprule
 & Criterion & Codes & Target \\
\midrule
C1 & Is the marked answer medically correct for this case? & 1 correct, 0 incorrect, ? cannot tell & 1 \\
C2 & Is the slide needed to answer? & V slide needed, T text alone, U not answerable from the slide & V \\
C3 & Is the question clinically meaningful and unambiguous? & Y yes, m minor issue, N no & Y \\
C4 & Are the wrong options plausible but wrong? & G all good, 1 one flawed, M several flawed & G \\
C5 & Does the stem give away the answer? & Y no leak, p partial hint, N leaks the answer & Y \\
C6 & Is the wording natural English? & Y natural, a awkward, N broken & Y \\
INC & Keep the question in the benchmark? & Y keep, N drop & Y \\
\bottomrule
\end{tabular}\end{adjustbox}
\end{table}

\begin{table}[t]\centering\footnotesize\setlength{\tabcolsep}{5pt}
\caption{Pathologist quality scores (\%). Each score estimates the share of
imported questions that meet the criterion. Task families are weighted by
their share of the imported tier, and brackets give stratified bootstrap
95\% CIs. $^{*}$No reviewed question failed this criterion. Morphologic
families are morphology, microscopy and diagnosis, which hold 83.90\% of
the imported questions.}
\label{tab:patho-scores}
\begin{tabular}{llcc}
\toprule
 & Criterion & All imported & Morphologic families \\
\midrule
\multicolumn{4}{l}{\textit{Text-level integrity}} \\
C5 & Stem does not give the answer away & 100.00$^{*}$ & 100.00$^{*}$ \\
C5 & No partial hint either & 98.10 [96.09--99.83] & 99.70 [99.10--100.00] \\
C6 & Wording is not broken & 100.00$^{*}$ & 100.00$^{*}$ \\
C6 & Wording is natural & 99.71 [99.20--100.00] & 99.70 [99.10--100.00] \\
C4 & All distractors plausible but wrong & 96.18 [92.70--98.84] & 100.00$^{*}$ \\
\midrule
\multicolumn{4}{l}{\textit{Clinical validity}} \\
C3 & Clinically meaningful and unambiguous & 91.69 [84.37--97.16] & 96.35 [89.04--100.00] \\
C2 & Not answerable from the text alone & 90.04 [71.67--99.91] & 89.67 [69.00--100.00] \\
C1 & Answer key medically correct & 85.60 [76.93--93.84] & 88.44 [78.08--96.35] \\
C2 & Slide needed to answer & 82.31 [63.29--95.51] & 86.01 [64.72--100.00] \\
\midrule
INC & Kept in the benchmark & 79.47 [60.21--92.77] & 86.01 [61.69--100.00] \\
\bottomrule
\end{tabular}
\end{table}

\section{Natural-image transfer}\label{app:transfer}

We apply the same loss and $\lambda$ to eight
vision--language models from five families: Qwen3-VL at 4B and 8B
\citep{qwen3vl}, InternVL3.5-8B \citep{internvl}, Qwen2.5-VL-7B
\citep{qwen25}, Qwen2-VL at 2B and 7B \citep{qwen2vl}, LLaVA-1.5-7B
\citep{llava15} and LLaVA-1.6-Mistral-7B \citep{llavanext}. A NaturalBench
item \citep{naturalbench} is the natural-image analogue of a counterfactual
slide pair: one human-verified question on two images whose correct answers
differ, designed so that it cannot be answered from text alone. We train on
1,886 NaturalBench pairs, of which 1,206 are yes-or-no and 680 are
two-option questions, and validate on 200 pairs. Because the released split
shares 406 images between training and test, we keep the 1,182 test pairs
whose two images never appear in training, as we did for SlideBench-TCGA.
Two questions on the same two images form the official group unit, and the
decontaminated test contains 585 complete groups.

For every base the adapter is a rank-16 LoRA on the attention projections
of the language model. Stage two runs 1,000 supervised steps on the 3,772
single items and stage three 800 further steps from that checkpoint. The CE
control uses $\lambda{=}0$, and pair-DPO uses $\lambda{=}5$ with the real
counterfactual image. Three alternative stage-three objectives use the same
data, schedule and number of updates: a retrieved similar image as the
negative \citep{realign}, symmetric ranking of the two images' answers
\citep{symmpo}, and anchoring on the image-free input \citep{svco}. G-acc is
the official group accuracy, counting a group only if all four answers are
correct, and Acc is per-question accuracy. Since the two images of a pair
have opposite answers, a model that ignores the image scores 50 on Acc and
0 on G-acc.

Table~\ref{tab:transfer} shows that most of the gain over zero-shot comes
from fine-tuning on the pairs itself: the CE control raises G-acc by 5.07
to 9.97 points, and pair-DPO adds 0.06 to 4.28 points, most on the two
LLaVA backbones. Pair-DPO has the highest paired sensitivity on all eight
backbones, tied on Qwen3-VL-4B, and exceeds the CE control by 6\% to 25\%,
with non-overlapping intervals on six. Its G-acc is best on five backbones
and second on the other three, though its interval overlaps those of all
other trained arms. The retrieved-image negative, whose answer need not
differ, is generally the weakest trained arm: it trails the CE control in
G-acc on seven backbones and falls below zero-shot sensitivity on two.
Pair-DPO thus transfers beyond pathology mainly as a gain in sensitivity,
with a small accuracy advantage over the strongest alternatives.

\clearpage

\setlength{\LTcapwidth}{\linewidth}

\begingroup
\small
\setlength{\tabcolsep}{2.5pt}
\renewcommand{\arraystretch}{0.9}

\begin{longtable}{@{}
>{\footnotesize\raggedright\arraybackslash}p{0.345\linewidth}
>{\footnotesize\centering\arraybackslash}p{0.205\linewidth}
>{\footnotesize\centering\arraybackslash}p{0.205\linewidth}
>{\footnotesize\centering\arraybackslash}p{0.205\linewidth}
@{}}

\caption{Natural-image transfer on the decontaminated NaturalBench test
(1,182 pairs, 585 complete groups), with group-clustered bootstrap 95\%
CIs on every number. G-acc is the official group accuracy, which counts a
group only when all four of its answers are correct, Acc is the
per-question accuracy over both images and Sens is the paired sensitivity
of Section~\ref{sec:cs-metrics}. The CE control and pair-DPO pool three
seeds and the other arms are single seeds. All trained arms see the same
data, the same schedule and the same number of updates. For each backbone,
\textbf{bold} marks the best and \underline{underline} the second-best
trained arm.}
\label{tab:transfer}\\

\toprule
Arm & G-acc & Acc & Sens \\
\midrule
\endfirsthead

\multicolumn{4}{@{}l}{\footnotesize\textit{Table \thetable, continued}} \\
\toprule
Arm & G-acc & Acc & Sens \\
\midrule
\endhead

\midrule
\multicolumn{4}{@{}l}{\footnotesize\textit{Continued on next page}} \\
\endfoot

\bottomrule
\endlastfoot

\multicolumn{4}{@{}l}{\footnotesize\textit{Qwen3-VL-8B}} \\*

zero-shot
& 30.26 [26.50--33.85]
& 77.71 [76.28--79.13]
& 1.129 [1.079--1.179] \\*

$+$CE control
& 38.86 [35.50--42.22]
& 80.61 [79.29--81.86]
& 1.184 [1.145--1.222] \\*

$+$retrieved image \citep{realign}
& 37.44 [33.50--41.37]
& 80.37 [79.00--81.77]
& 1.161 [1.119--1.204] \\*

$+$sym. response \citep{symmpo}
& 37.61 [33.50--41.71]
& 80.54 [79.06--81.94]
& \underline{1.260 [1.217--1.302]} \\*

$+$blind-anchored \citep{svco}
& \textbf{41.03 [36.92--44.96]}
& \textbf{81.56 [80.03--82.97]}
& 1.239 [1.198--1.281] \\*

$+$pair-DPO (Ours)
& \underline{39.83 [36.07--43.76]}
& \underline{81.01 [79.56--82.47]}
& \textbf{1.277 [1.232--1.321]} \\

\midrule

\multicolumn{4}{@{}l}{\footnotesize\textit{Qwen3-VL-4B}} \\*

zero-shot
& 28.72 [25.13--32.31]
& 77.28 [75.91--78.70]
& 1.140 [1.094--1.191] \\*

$+$CE control
& \underline{37.61 [34.19--41.08]}
& 80.08 [78.80--81.35]
& 1.176 [1.135--1.214] \\*

$+$retrieved image \citep{realign}
& 33.16 [29.40--36.92]
& 78.55 [77.09--79.98]
& 1.131 [1.087--1.173] \\*

$+$sym. response \citep{symmpo}
& 36.75 [32.65--40.68]
& 79.86 [78.40--81.29]
& \underline{1.231 [1.183--1.280]} \\*

$+$blind-anchored \citep{svco}
& \underline{37.61 [33.50--41.54]}
& \textbf{80.46 [79.03--81.96]}
& \textbf{1.247 [1.201--1.293]} \\*

$+$pair-DPO (Ours)
& \textbf{37.78 [33.68--41.71]}
& \underline{80.25 [78.83--81.69]}
& \textbf{1.247 [1.198--1.292]} \\

\midrule

\multicolumn{4}{@{}l}{\footnotesize\textit{InternVL3.5-8B}} \\*

zero-shot
& 31.11 [27.35--34.70]
& 76.61 [75.02--78.16]
& 1.081 [1.036--1.126] \\*

$+$CE control
& \underline{37.38 [33.73--40.63]}
& 79.85 [78.54--81.15]
& 1.135 [1.095--1.173] \\*

$+$retrieved image \citep{realign}
& 34.70 [30.94--38.63]
& 78.93 [77.44--80.52]
& 1.066 [1.028--1.104] \\*

$+$sym. response \citep{symmpo}
& 36.07 [32.31--40.00]
& 79.57 [78.13--81.03]
& 1.174 [1.132--1.217] \\*

$+$blind-anchored \citep{svco}
& 36.75 [32.99--40.51]
& \underline{80.03 [78.61--81.41]}
& \underline{1.184 [1.141--1.229]} \\*

$+$pair-DPO (Ours)
& \textbf{37.44 [33.50--41.37]}
& \textbf{80.25 [78.86--81.69]}
& \textbf{1.241 [1.195--1.289]} \\

\midrule

\multicolumn{4}{@{}l}{\footnotesize\textit{Qwen2.5-VL-7B}} \\*

zero-shot
& 25.30 [22.05--28.89]
& 75.34 [73.83--76.80]
& 0.983 [0.946--1.019] \\*

$+$CE control
& \underline{32.42 [29.12--35.78]}
& 77.79 [76.38--79.21]
& 1.074 [1.035--1.112] \\*

$+$retrieved image \citep{realign}
& 30.09 [26.32--33.85]
& 77.16 [75.66--78.66]
& 1.095 [1.053--1.137] \\*

$+$sym. response \citep{symmpo}
& 31.79 [28.03--35.38]
& \underline{78.09 [76.58--79.55]}
& 1.116 [1.077--1.157] \\*

$+$blind-anchored \citep{svco}
& 31.28 [27.52--35.05]
& 77.50 [75.97--79.05]
& \underline{1.122 [1.079--1.164]} \\*

$+$pair-DPO (Ours)
& \textbf{33.16 [29.40--36.92]}
& \textbf{78.30 [76.83--79.76]}
& \textbf{1.222 [1.173--1.269]} \\

\midrule

\multicolumn{4}{@{}l}{\footnotesize\textit{Qwen2-VL-7B}} \\*

zero-shot
& 24.62 [21.19--28.21]
& 73.82 [72.26--75.32]
& 0.888 [0.852--0.925] \\*

$+$CE control
& 32.88 [29.46--36.13]
& 78.06 [76.68--79.45]
& \underline{1.053 [1.013--1.092]} \\*

$+$retrieved image \citep{realign}
& 31.28 [27.35--34.87]
& 76.44 [74.85--78.02]
& 1.019 [0.981--1.060] \\*

$+$sym. response \citep{symmpo}
& 31.97 [28.03--35.73]
& 77.58 [75.99--79.07]
& 1.037 [0.999--1.076] \\*

$+$blind-anchored \citep{svco}
& \textbf{34.53 [30.77--38.29]}
& \underline{78.09 [76.52--79.61]}
& 1.049 [1.012--1.089] \\*

$+$pair-DPO (Ours)
& \underline{34.30 [30.83--37.49]}
& \textbf{78.24 [76.82--79.63]}
& \textbf{1.115 [1.075--1.154]} \\

\midrule

\multicolumn{4}{@{}l}{\footnotesize\textit{Qwen2-VL-2B}} \\*

zero-shot
& 18.46 [15.21--21.37]
& 71.19 [69.68--72.67]
& 0.737 [0.702--0.771] \\*

$+$CE control
& 23.53 [20.57--26.78]
& 73.65 [72.32--74.97]
& 0.820 [0.784--0.856] \\*

$+$retrieved image \citep{realign}
& 20.85 [17.61--24.10]
& 72.34 [70.81--73.88]
& 0.818 [0.781--0.855] \\*

$+$sym. response \citep{symmpo}
& \textbf{25.81 [22.39--29.40]}
& \textbf{74.58 [72.97--76.10]}
& \underline{0.908 [0.869--0.943]} \\*

$+$blind-anchored \citep{svco}
& 24.44 [21.02--27.87]
& \underline{74.37 [72.89--75.87]}
& 0.890 [0.851--0.930] \\*

$+$pair-DPO (Ours)
& \underline{25.64 [22.45--28.95]}
& 74.10 [72.64--75.51]
& \textbf{0.918 [0.879--0.958]} \\

\midrule

\multicolumn{4}{@{}l}{\footnotesize\textit{LLaVA-1.5-7B}} \\*

zero-shot
& 8.38 [6.15--10.77]
& 61.84 [60.45--63.32]
& 0.430 [0.402--0.456] \\*

$+$CE control
& 18.35 [15.90--20.97]
& 69.54 [68.23--70.84]
& 0.728 [0.695--0.764] \\*

$+$retrieved image \citep{realign}
& 19.32 [16.07--22.91]
& 69.63 [68.01--71.08]
& 0.694 [0.659--0.730] \\*

$+$sym. response \citep{symmpo}
& 18.29 [15.04--21.71]
& 68.57 [66.91--70.23]
& 0.713 [0.677--0.750] \\*

$+$blind-anchored \citep{svco}
& \underline{19.49 [16.24--22.91]}
& \underline{70.14 [68.57--71.75]}
& \underline{0.742 [0.708--0.779]} \\*

$+$pair-DPO (Ours)
& \textbf{21.03 [17.78--24.28]}
& \textbf{71.19 [69.68--72.81]}
& \textbf{0.908 [0.863--0.951]} \\

\midrule

\multicolumn{4}{@{}l}{\footnotesize\textit{LLaVA-1.6-Mistral-7B}} \\*

zero-shot
& 14.53 [11.79--17.44]
& 69.25 [67.80--70.68]
& 0.710 [0.677--0.744] \\*

$+$CE control
& \underline{24.10 [21.20--27.18]}
& 74.11 [72.82--75.36]
& 0.895 [0.859--0.932] \\*

$+$retrieved image \citep{realign}
& 23.25 [20.17--26.67]
& 73.60 [72.01--75.11]
& 0.824 [0.790--0.860] \\*

$+$sym. response \citep{symmpo}
& 22.91 [19.49--26.50]
& 73.98 [72.54--75.49]
& \underline{0.938 [0.897--0.977]} \\*

$+$blind-anchored \citep{svco}
& 23.93 [20.34--27.18]
& \underline{74.24 [72.80--75.67]}
& 0.905 [0.866--0.942] \\*

$+$pair-DPO (Ours)
& \textbf{28.38 [24.96--32.14]}
& \textbf{75.42 [73.86--76.93]}
& \textbf{1.062 [1.015--1.109]} \\

\end{longtable}

\endgroup

\end{document}